\documentclass[10pt, a4paper]{article}
\usepackage{lrec2022} 
\usepackage{multibib}
\newcites{languageresource}{Language Resources}
\usepackage{graphicx}
\usepackage{tabularx}
\usepackage{soul}
\usepackage[utf8]{inputenc}
\usepackage[T2A,LAE,T1]{fontenc}
\usepackage[arabic,USenglish]{babel}
\usepackage{mathtools} 

\usepackage{titlesec}
\titleformat{\section}{\normalfont\large\bfseries\center}{\thesection.}{1em}{}
\titleformat{\subsection}{\normalfont\SmallTitleFont\bfseries\raggedright}{\thesubsection.}{1em}{}
\titleformat{\subsubsection}{\normalfont\normalsize\bfseries\raggedright}{\thesubsubsection.}{1em}{}
\renewcommand\thesection{\arabic{section}}
\renewcommand\thesubsection{\thesection.\arabic{subsection}}
\renewcommand\thesubsubsection{\thesubsection.\arabic{subsubsection}}
\usepackage{array, makecell} %
\usepackage{epstopdf}
\usepackage[utf8]{inputenc}

\usepackage{hyperref}
\usepackage{xstring}

\usepackage{color}

\usepackage{xcolor} 
\definecolor{LightGray}{gray}{0.9}

\title{GLARE: Google Apps  Arabic Reviews Dataset\\ \vspace*{.5\baselineskip}}

\name{Fatima AlGhamdi\textsuperscript{1}, Reem Mohammed\textsuperscript{1}, Hend Al-Khalifa\textsuperscript{2}, and Areeb Alowisheq\textsuperscript{1,3}}

\address{\small \textsuperscript{1} National Center for Artificial Intelligence (NCAI), Saudi Data and Artificial Intelligence Authority (SDAIA), Riyadh 12391, Saudi Arabia\\ 
\small \textsuperscript{2} College of Computer and Information Sciences, King Saud University, Riyadh 11495, Saudi Arabia\\ 
\small \textsuperscript{3} College of Computer and Information Sciences, Imam Mohammad Ibn Saud Islamic University, Riyadh 13318, Saudi Arabia\\
\texttt{\{falghamdi, rmohammed\}@sdaia.gov.sa}, \texttt{hendk@ksu.edu.sa}, \texttt{aaalowisheq@imamu.edu.sa}}

\abstract{
This paper introduces GLARE an Arabic Apps Reviews dataset collected from Saudi Google PlayStore. It consists of 76M reviews, 69M of which are Arabic reviews of 9,980 Android Applications. We present the data collection methodology, along with a detailed Exploratory Data Analysis (EDA) and Feature Engineering on the gathered reviews. We also highlight possible use cases and benefits of the dataset.
 \\ \newline  }

\begin{document}

\maketitleabstract

\section{Introduction}

Various Natural Language Processing (NLP) tasks such as: Topic Modeling, Sentiment Analysis (SA) and Aspect Based Sentiment Analysis (ABSA), require good quality datasets  \cite{nassif2021deep}. But there are issues that lie behind the availability of such data when it comes to low resource languages like the Arabic language \cite{alsarsour2018dart}. 

\medbreak

The Arabic language is a complex and rich language. It has different forms depending on the region (dialects) or usage. It can be classified into: Dialectal Arabic (DA), Modern Standard Arabic (MSA) and Classical Arabic (CA) \cite{abdul2018you}. Dealing with DA can be challenging compared to MSA, as the former has a shortfall of different NLP tools and resources built to support it \cite{al2019comprehensive}. Nevertheless, the language itself lacks more diverse resources since many of the recent available datasets are from social media platforms, especially Twitter \cite{nassif2021deep}.

\medbreak
 
In this work, we present Google App Arabic Reviews dataset (GLARE). A dataset of Android Applications reviews collected from the Saudi Google PlayStore \footnote{\url{https://play.google.com/store/}}. It differs from other available datasets in size, and to best of our knowledge, it is the largest Arabic reviews dataset to date, with 76M reviews covering 9,980 Android Applications. Although, similar datasets such as \cite{al2022designing} exist, yet with 51K reviews its significantly smaller than GLARE.
We believe that GLARE will be a good contribution to the Arabic NLP community for two reasons: its huge size and its domain. Compared to tweets, the character limit for App reviews on Google play is longer, as up to 4000 characters are permitted compared to 280 characters on Twitter, this allows for more expressive sentences. Thus, the dataset will be helpful to many researchers who are looking to utilize it for NLP tasks such as Sentiment Analysis or Aspect Based Sentiment Analysis, as a good volume of data is needed to perform any Artificial Intelligence (AI) related practice \cite{ahmed2022freely}  \medbreak
The rest of the paper is organized as follows: Section 2  reviews related work in the domain of app reviews, Section 3  presents our data collection methodology, Section 4  analyses GLARE dataset, Section 5 further explores GLARE dataset by applying feature engineering to extract additional features, Section 6 highlights possible use cases and benefits of the dataset. Finally, Section 7 concludes the paper with future work. 

\section{Related Work} 
Researchers are continuously contributing high-quality resources for the Arabic NLP community. Masader \cite{alyafeai2021masader} provides a public catalogue for over 200 Arabic NLP datasets. The majority of these datasets were from social media platforms, predominantly from Twitter. 
\\
\\
As examples of Twitter datasets, \cite{haouari2020arcov} published the first Arabic tweets dataset related to COVID-19. They collected approximately 2.7M tweets using Twitter search API. Similarly, \cite{mulki2021let} released a dataset targeting the "Levantine" Arabic dialect. The dataset is intended to be used for the misogyny language detection task. Another work by \cite{alharbi2020asad} where they published a benchmark dataset that consists of 95k annotated tweets with sentiment labels of multiple Arabic dialects.
\\
\\
As for other types of datasets, \cite{einea2019sanad} aimed to provide an Arabic dataset that can be used for text classification/categorization tasks. The dataset consisted of Arabic news articles gathered from different news portals, it consists of 200K articles distributed between 7 categories.  \cite{elnagar2018hotel} published an Arabic hotel reviews dataset. The dataset was collected to be used for SA tasks and it had approximately 38K reviews covering 1,858 hotels. Likewise, \cite{aly2013labr} collected over 63K Arabic book reviews from the website Goodreads \footnote{\url{https://www.goodreads.com/}} representing 2,143 books. Another related work was by \cite{al2015human} where they published a benchmark dataset using a subset of book reviews from the previously mentioned work but in contrast, it was manually annotated with the intention for it to be used for ABSA task, the dataset contained 1,513 reviews. A final example, \cite{ali2021arafacts} presented an Arabic dataset that could be used for fact-checking. They crawled 6,222 claims from 5 Arabic fact-checking websites and were shared to the public.
\\
\\
Given the previous work in Arabic dataset creation, our intention in this work is to release a dataset that is not only large in size but also from an under-represented source for Arabic data such as App Stores Reviews. We believe that our dataset can benefit both the NLP and Software Development communities. 

\section{Data Collection Methodology}

\subsection{Approach}

GLARE dataset was harvested from Google PlayStore using google-play-scraper\footnote{\url{https://github.com/facundoolano/google-play-scraper}\label{google-scraper}} library in Nodejs for crawling and its python\footnote{\url{https://github.com/JoMingyu/google-play-scraper}} version was used for scraping reviews and their metadata. We scraped reviews from top 200 free apps from each main and sub category in the Saudi Google PlayStore, which resulted to a total of 59 categories and over 11K apps. We chose to scrape free apps since they are accessible to all and; hence, should have more reviews than paid apps. After dropping duplicated apps, we ended up with a total of 9,980 unique apps. The number of retrieved reviews is over 76M reviews with a total size of 17 Gigabytes (including apps metadata). After applying pre-processing steps, which include dropping duplicates and null review content, removing symbols, numbers and noise, and keeping only Arabic reviews,  we ended up with over 69M Arabic app reviews. The raw and engineered \ref{FeatureEngineering} datasets are described in Table \ref{tab:glare-data} and available for download via GitHub\footnote{\url{https://github.com/Fatima-Gh/GLARE}} and Hugging Face\footnote{\url{https://huggingface.co/datasets/Fatima-Gh/GLARE}}. A summary of GLARE dataset is presented in Table \ref{tab:data-collection-stat}.\\ 

\begin{table}[!h]
\setcellgapes{3pt}\makegapedcells
\begin{center}
\begin{tabularx}{\columnwidth}{|X|X|X|}
\hline
 \textbf{Data Type }& \textbf{File Name}& \textbf{File Size}\\ \hline
 raw&  apps&4.1 MB\\ \hline
 raw& reviews&17 GB \\ \hline
 raw&  categories &4.3 MB \\ \hline
 engineered&  apps& 3.8 MB\\ \hline
 engineered&  reviews& 21.9 GB\\ \hline
 engineered&  vocabulary& 530.5 MB\\ \hline
\end{tabularx}
\caption{An Overview of GLARE Raw and Engineered Data.}
\label{tab:glare-data}
\end{center}
\end{table}

\begin{table}[!h]
\begin{center}
\begin{tabularx}{\columnwidth}{|X|l|l|l|X|}
\hline
 \textbf{Store} & \textbf{Apps} & \textbf{Reviews} &  \textbf{Size} & \textbf{Period} \\
 \hline
 Google PlayStore &  9,980 &  76M &  17 GB & March 21 - April 21  \\ 
 \hline
\end{tabularx}
\caption{Statistics of Collected Data.} 

\label{tab:data-collection-stat}
\end{center}
\end{table}

\subsection{Apps Metadata}

\begin{figure}[htbp]
    \centering
    \includegraphics[width=\linewidth]{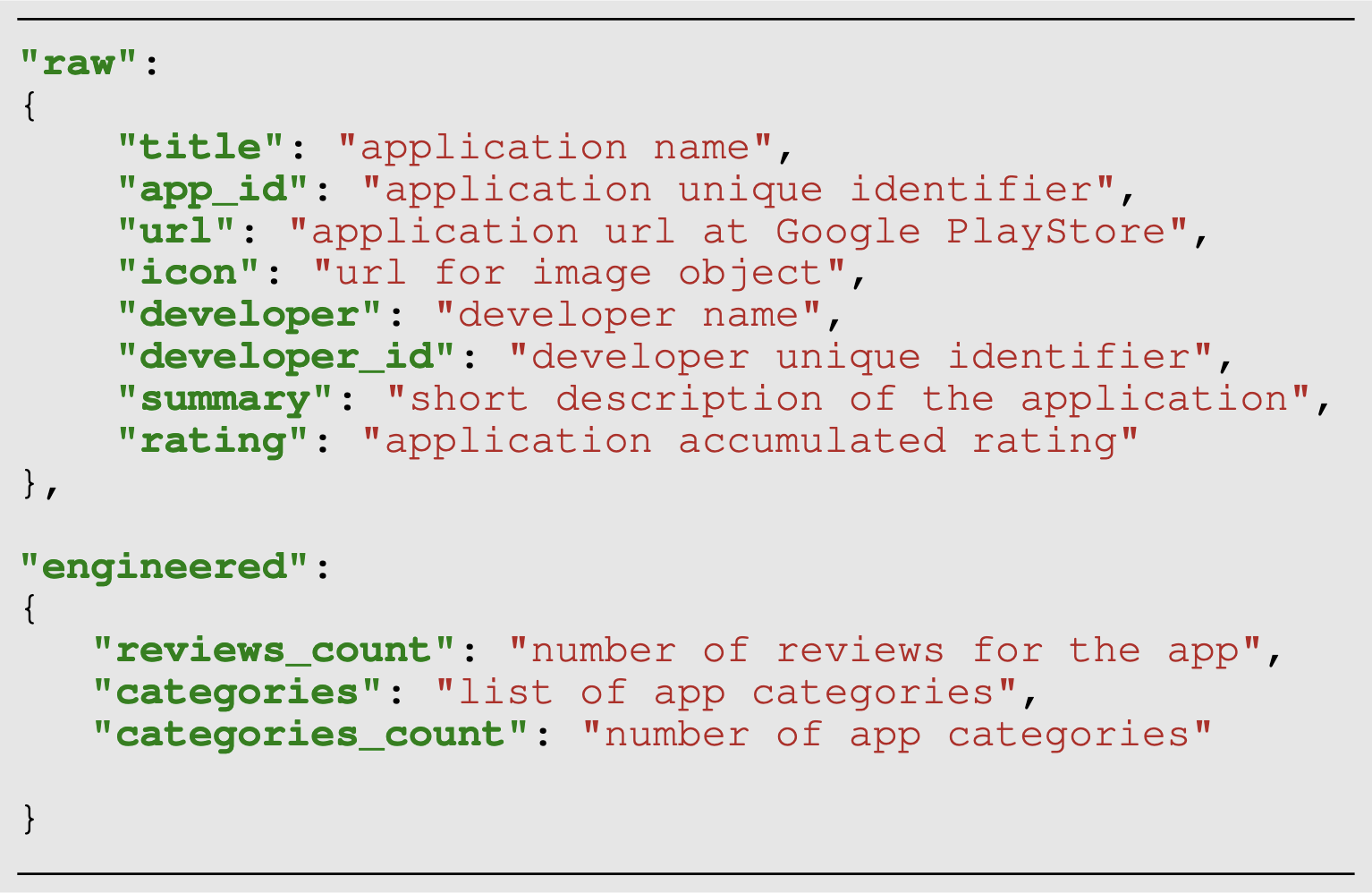}
    \caption{Apps Metadata}
    \label{apps-metadata}
\end{figure}

Metadata extracted from Google PlayStore holds useful information about the application that can be used to derive insights using statistical analysis and machine learning approaches. These data include app rating (score) at the time of data scraping, application ID and URL in the PlayStore, icon image URL, and app summary. An overview of the collected and engineered metadata is presented in Figure \ref{apps-metadata}.

\begin{figure}[htbp]
    \centering
    \includegraphics[width=\linewidth]{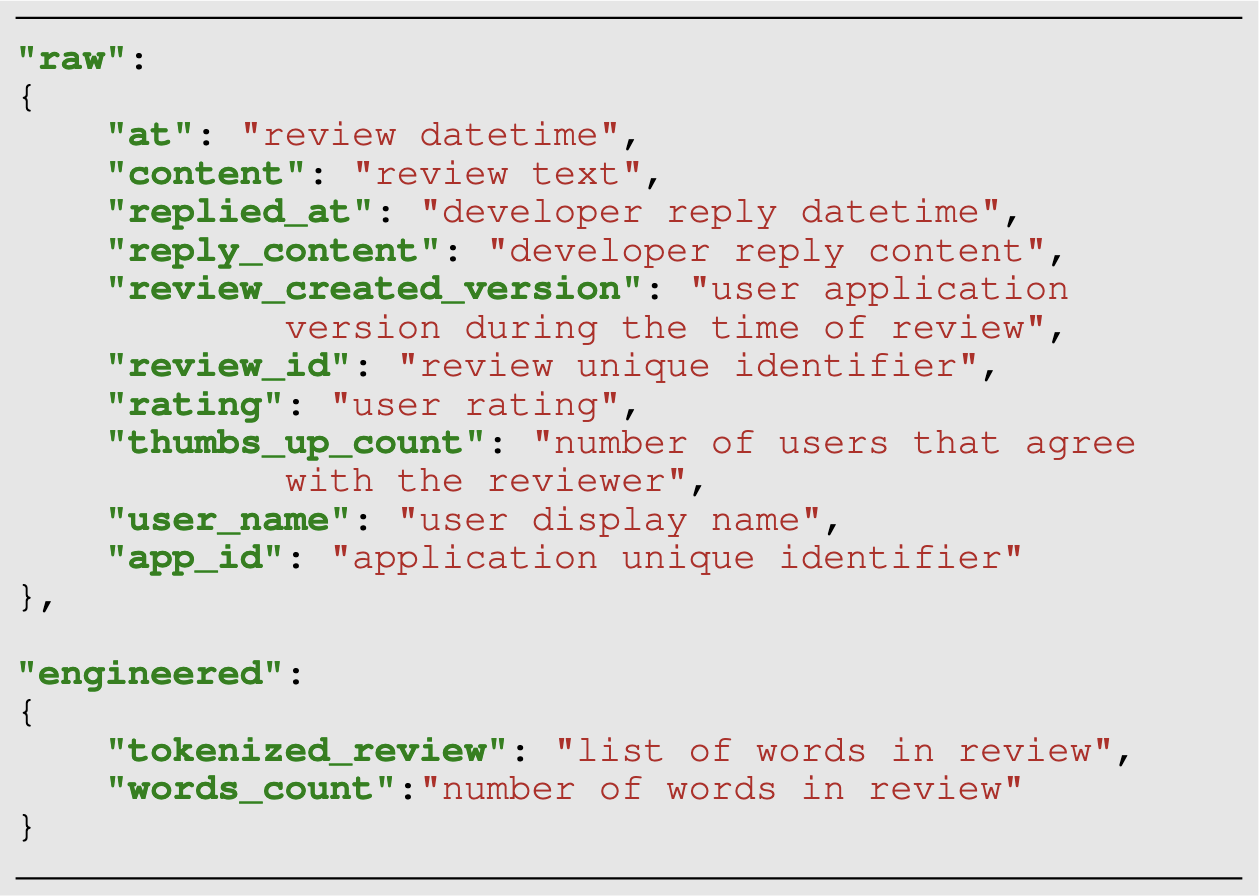}
    \caption{Reviews Metadata} 
    \label{reviews-metadata}
\end{figure}
 






\subsection{Reviews Metadata}

The reviews metadata also offers valuable information such as user rating associated with the review, number of users that agree with the reviewer, user display name and review app version. This  data is extracted and published along with the reviews. A more descriptive overview of the raw and engineered reviews metadata can be found in Figure \ref{reviews-metadata}.

\section{Dataset Analysis} 

To understand the properties of GLARE dataset and the insights that can be derived from the collected data, we conduct the following descriptive analysis:
\begin{itemize}
     \item {The distribution of reviews ratings and the number of users that agree with the reviewer.}
    \item{Developers engagement with users.}
 \end{itemize}

\subsection{Ratings and Thumbs-up Count Distribution}
\subsubsection{Ratings}

To write a review in Google PlayStore, it is mandatory for the user to provide a rating or score of the application that ranges from 1 to 5. Analysing ratings of the reviews and their effect on various properties such as inciting developers to reply or how user ratings change overtime can be of help in software maintenance and evolution life-cycle \cite{dkabrowski2022analysing}. In our dataset, we found that the ratings are skewed greatly with over 80\% of the reviews having 5 stars. Additional statistics of ratings distribution is presented in Figure \ref{fig.3}.\\

\subsubsection{Thumbs-up}
Google PlayStore provides user-to-user engagement functionality through a voting mechanism. Any user can view and up-vote in agreement with other users' reviews. This feature provides useful insights of the general sentiments of an application's customer population. Over 98\% of the apps had reviews with thumbs-ups, with a total of 8.1M reviews. The highest thumbs-up for a review was 67K while the lowest was of 1.

\subsubsection{Thumbs-up and Ratings Distribution}

\begin{figure}[h]
\begin{center}
\includegraphics[scale=0.25]{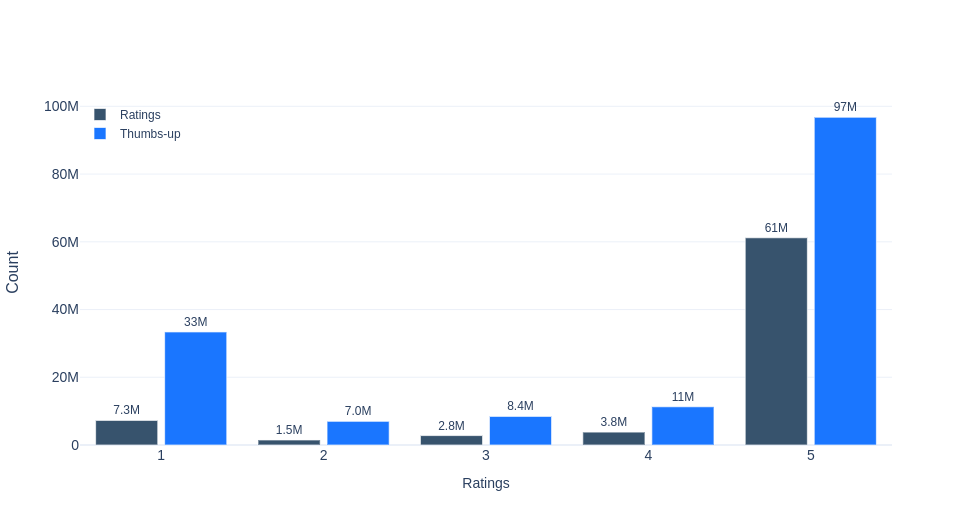} 

\caption{Statistics of Thumbs-up with respect to Ratings Distribution.}
\label{fig.3}
\end{center}
\end{figure}

To show a descriptive analysis of the distribution of the previously mentioned features, reviews' ratings were mapped with up-votes.  The results are shown in Figure \ref{fig.3}.\\

\subsection{Developers Reply to Users}

\begin{figure}[h]
\begin{center}

\includegraphics[scale=0.40]{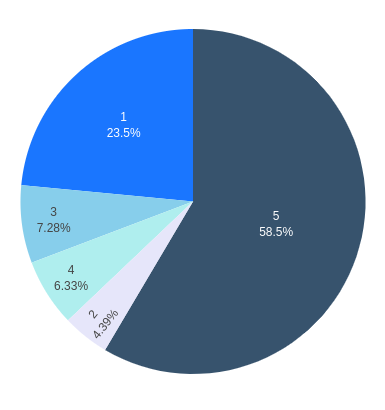} 

\caption{Percentage of Developers Engagement with respect
to Reviews Ratings (1 to 5).}
\label{fig.4}
\end{center}
\end{figure}

Developers engagement with the users is defined as developers replying to users' reviews in the app store. The affect of such interaction can provide valuable analysis on customer behavioral patterns when service providers engage with them. In GLARE dataset, about 48\% of apps had engaged with customer reviews with a total of 3.7M developers' reply. The highest developer engagement was by Azar App, a video chat and livestream application, with over 203K developers replies. The reviews ratings and developer engagement distribution is presented in Figure \ref{fig.4}.\\

\section{Feature Engineering}
To further explore our GLARE dataset, feature engineering methods were applied to extract additional features from the raw data that can be utilized for machine learning and NLP modeling tasks. Some of these methods include engineering a vocabulary dataset, identifying words count per review and reviews count per app and engineering duplicated apps. The results of the feature engineering process is shown in Table \ref{tab:engineered-features-stat}.
\label{FeatureEngineering}

\begin{table}[!h]
\setcellgapes{3pt}\makegapedcells
\begin{center}
\begin{tabularx}{\columnwidth}{|X|l|}

\hline
 \textbf{Raw and Engineered Features }& \textbf{Result}\\ \hline
 Total Number of Reviews&  76,512,077\\ \hline
 Total Number of Non-empty Reviews& 76,387,928\\ \hline
 Vocabulary Count &  8,707,776 \\ \hline
 Tokens with Term Frequency = 1&  6,960,449 \\ \hline
 Longest Review Sequence (Tokens Count)&  753 \\ \hline
 Shortest Review Sequence (Tokens Count)&  1 \\ \hline
 Number of Reviews with One Word&  30,953,303 \\ \hline
 Percentage of Reviews with One Word&  40.5\% \\ \hline
 Reviews Replied by Developers&  3,798,136 \\ \hline
 Reviews with 5-star Ratings&  61,184,348 \\ \hline
 Highest Number of Reviews per App&  3,207,530 \\ \hline
 Lowest Number of Reviews per App&  1 \\ \hline
 Number of Apps with One Review&  216 \\ \hline
 Longest Character Count for a Word&  2,383 \\ \hline
 Maximum Number of Categories per App& 4 \\ \hline
 Number of Main and Sub Categories & 59 \\ \hline
 Number of Categories Combinations& 213 \\ \hline
\end{tabularx}
\caption{A Summary of Raw and Engineered Features Statistics.}
\label{tab:engineered-features-stat}
\end{center}
\end{table}



\begin{figure*}[ht]
\begin{center}
  \begin{tabular}{@{}cc@{}}
    \multicolumn{2}{c}{\includegraphics[width=0.7\linewidth]{    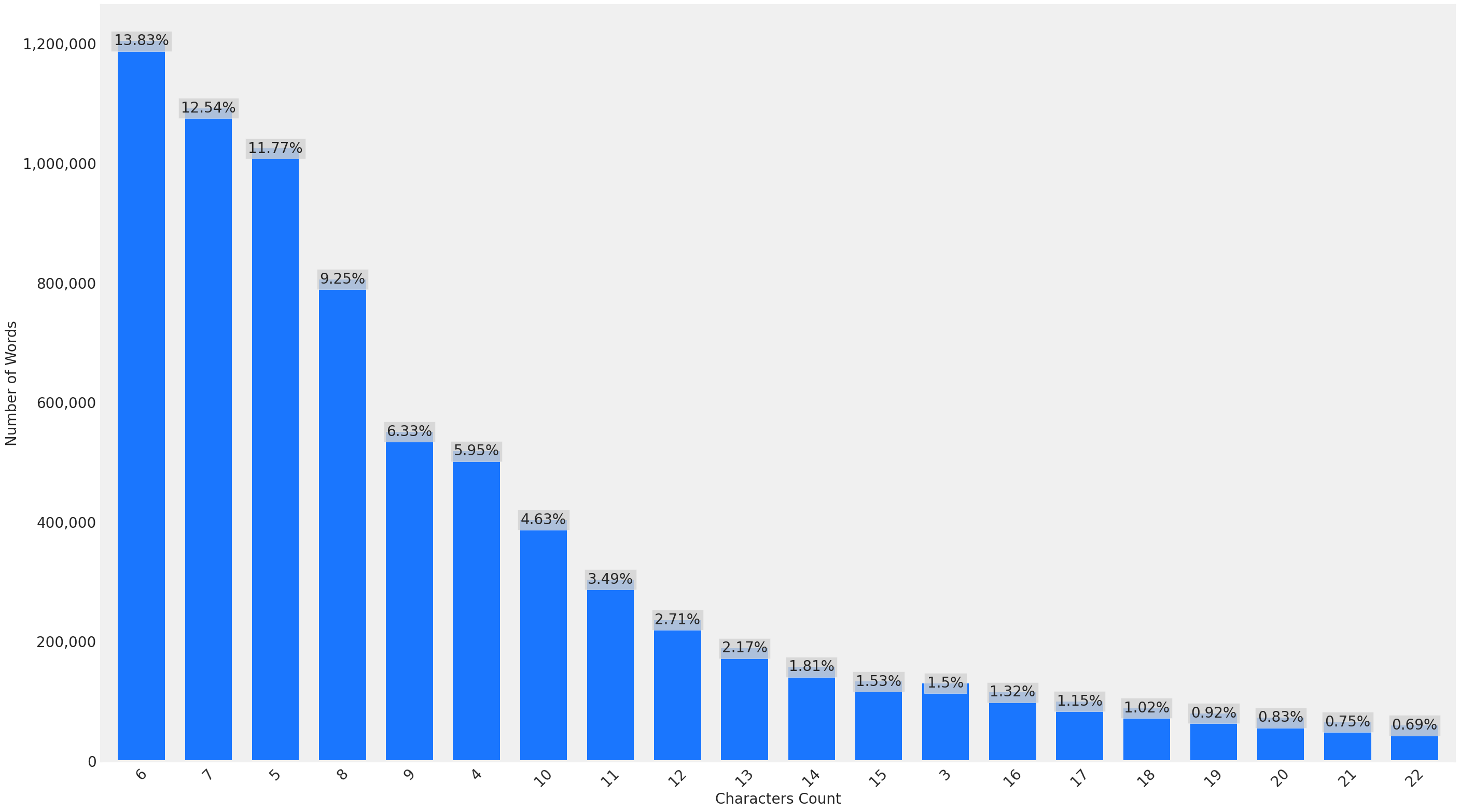} }
  \end{tabular}
\caption{Top 80\% Most Frequent Characters Length per Word with respect to the Total Number of Words in the Vocabulary Dataset.}
\label{char-length}
\end{center}
\end{figure*}

\begin{figure}[h]
\begin{center}

\includegraphics[scale=0.15]{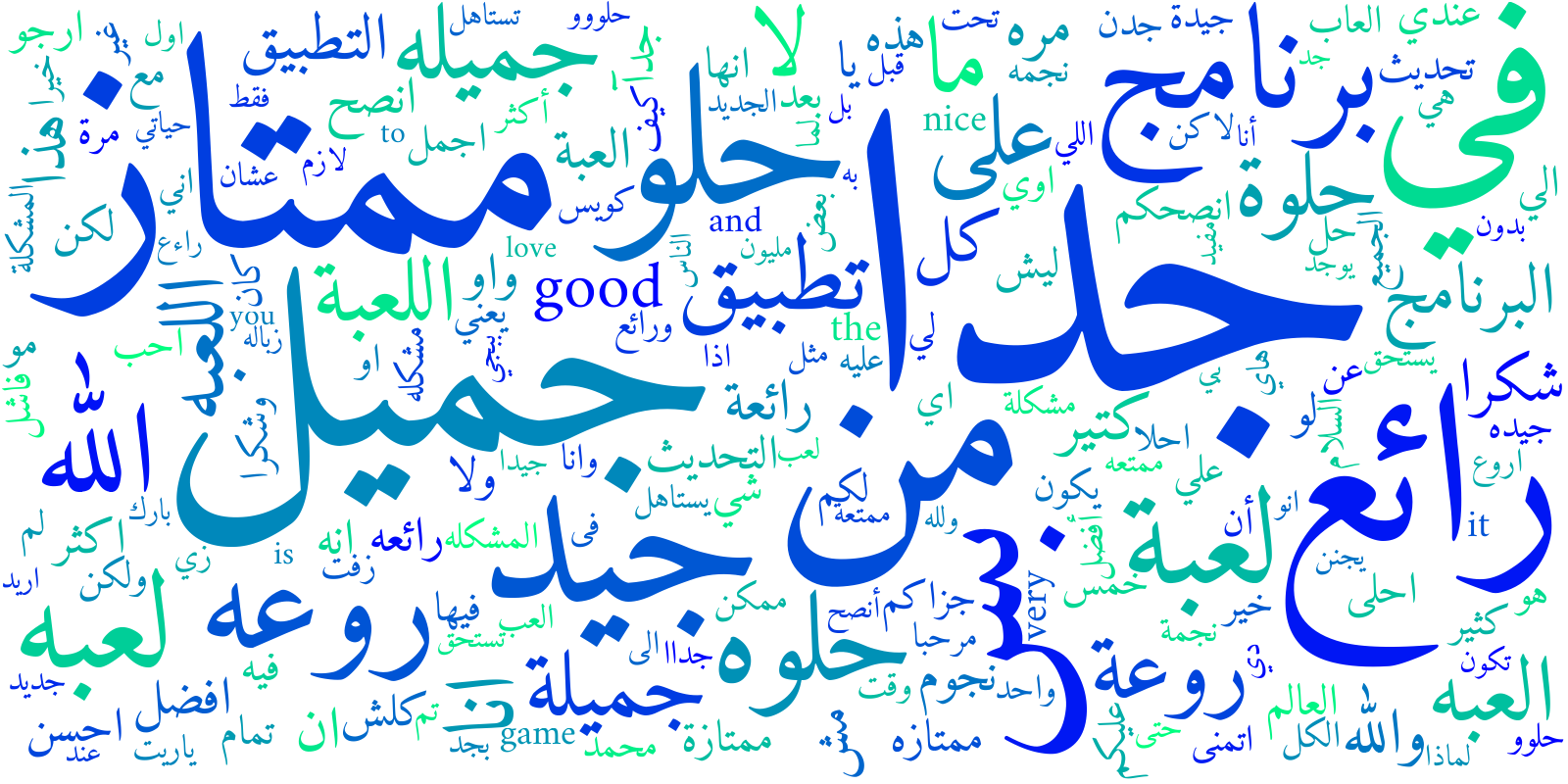}

\caption{Top 200 Most Frequent Words.}
\label{fig.7}
\end{center}
\end{figure}

\subsection{Term Dictionary}

To identify words' statistics and noise distribution in the dataset, we constructed term frequency dictionary using CountVectorizer \cite{scikit-learn} with its default settings that split words on white-space and punctuation and keep words with characters count greater or equal to 2. Features (words) and their occurrences are extracted from all the reviews in GLARE dataset to construct a vocabulary dictionary. The total number of unique words are 8.7M, with 6.9M words only appearing once in the reviews dataset. Noise in the vocabulary constitutes around 17\% of the words, while 15\% of the vocabulary words had digits; hence, we define noise as any character that is non-Arabic alphanumeric. We use this information to understand how much noise exist in the vocabulary set and what cleaning steps could be applied on GLARE dataset. The longest word based on character count was 2,383 character long, followed by 1,557, 1,362, 1,340, and 1,317 characters long. We found that all the words with very high character count had only appeared once in the whole dataset. 13.83\% of the words in the vocabulary have a length of 6 characters as shown in Figure \ref{char-length}. As for the top frequent terms, the word \AR{جدا} \emph{jidan} has the highest term frequency with 12M occurrences followed by \AR{ممتاز} \emph{mumtaz} with 5.6M occurrences and \AR{جميل} \emph{jameel} with 5M occurrences \ref{fig.7}. \\

\subsection{Tokenization and Review Length}

\begin{figure*}
\begin{center}
  \begin{tabular}{@{}cc@{}}
    \multicolumn{2}{c}{\includegraphics[width=0.8\linewidth]{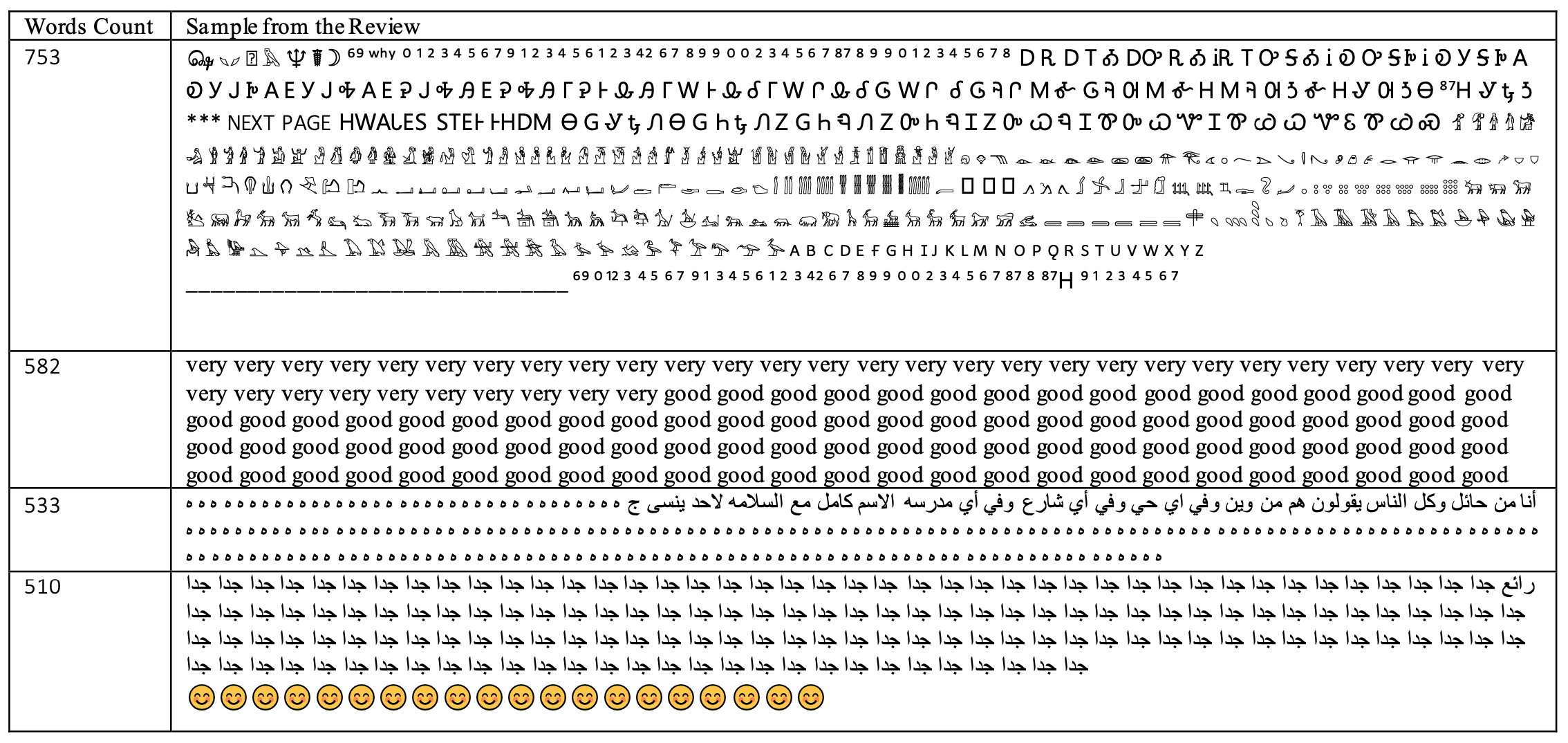}}
  \end{tabular}
\caption{Samples from the Top Longest Reviews.}
\label{fig.5}
\end{center}
\end{figure*}

To understand words per review distribution, the reviews were tokenized on white-space, and words count was calculated by getting the length of the tokenized review. The longest review based on words sequence was 753 words long, while the shortest review consist of 1 word. The top longest reviews mainly consisted of noise, which includes symbols, digits, and emojis, and repeated terms. More than 40\% of the reviews are of length 1, which in total amounts to over 30.9M reviews. Samples from the top longest reviews are shown in Figure \ref{fig.5}

\subsection{Duplicated Apps}

While collecting apps from the Saudi Google PlayStore, each app was mapped to a set of apps' categories, i.e. apps on PlayStore can have more than one category label. The number of extracted apps was 11,736 apps from 59 categories. The application ID was used to scrape reviews, and duplicated apps were only scraped once. Duplicated apps were handled by keeping only one instance of the app. The final list of selected applications consists of 9,980 unique apps, which was engineered by removing duplicated apps that had exactly the same values in all the properties. As for apps that had different ratings, we averaged the ratings after dropping duplicates with rating of 0 to avoid skewing the average, and then we kept only one instance of the app with the updated rating. 

\subsection{Reviews Count per App}

\begin{figure*}[ht]
\begin{center}
  \begin{tabular}{@{}cc@{}}
    \multicolumn{2}{c}{\includegraphics[width=0.8\linewidth]{    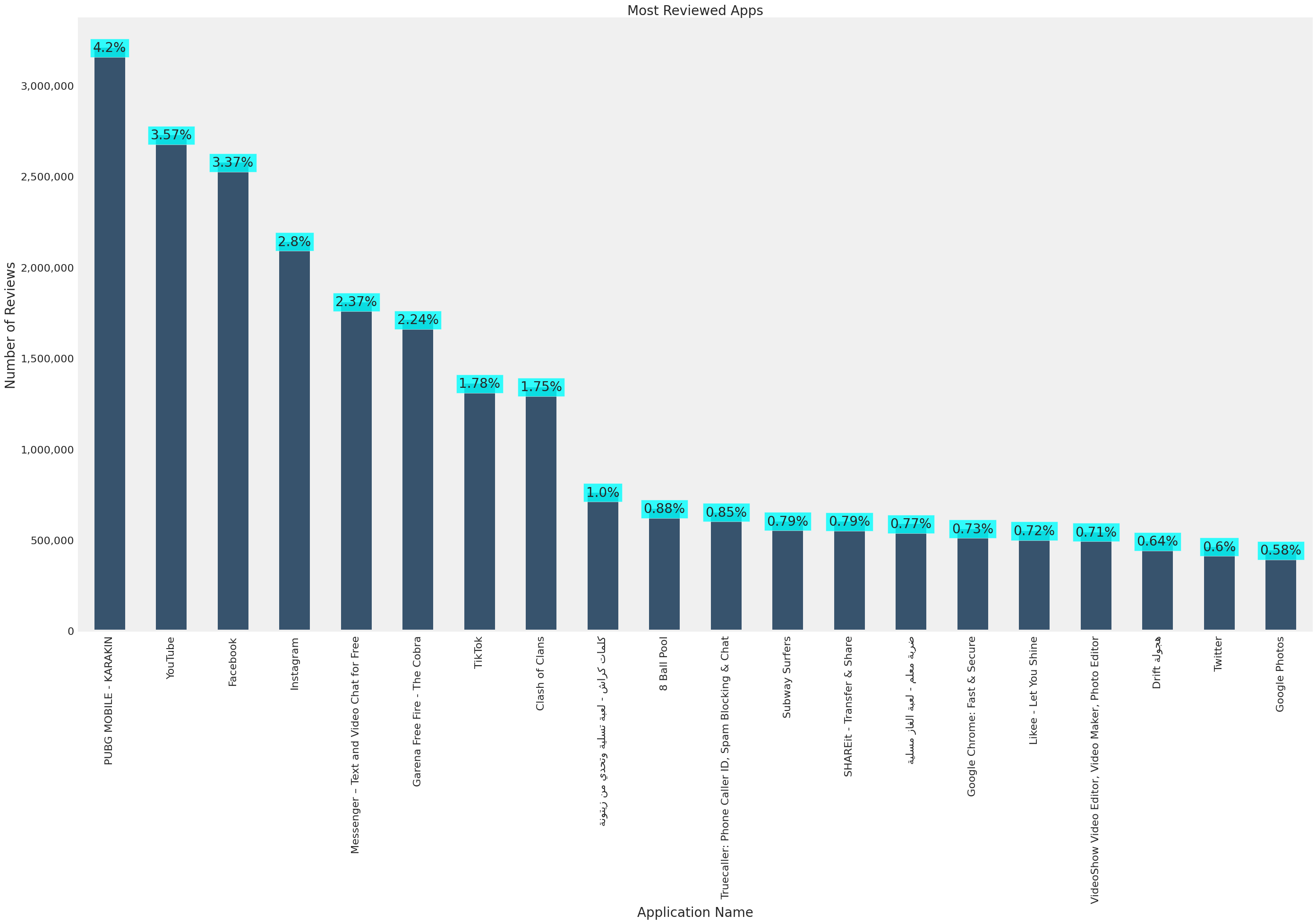} }
  \end{tabular}
\caption{Percentage of Reviews in Top 20 Most Reviewed Apps.}
\label{fig.6}
\end{center}
\end{figure*}

In GLARE dataset, the highest number of reviews for an app is 3.2M reviews for PUBG, followed by YouTube with 2.7M reviews, and Facebook with 2.5M reviews. While the lowest number of reviews for an app was 1 review.  2.2\% of the dataset apps had only one review, this equals to 216 apps. Additional information of top apps based on review count is presented in Figure \ref{fig.6}

\begin{table*}[ht]
\begin{center}
\begin{tabular}{|l|l|l|}
\hline
 \textbf{Category Combination}& \textbf{No. of Apps}& \textbf{No. of Reviews}\\ \hline
 (maps\_and\_navigation)&  190&274,901\\ \hline
 (family\_create, art\_and\_design)& 11&6,046 \\ \hline
 (android\_wear, health\_and\_fitness, application)&  1 &904 \\ \hline
 (game\_music, family, game, family\_musicvideo)&  1& 9,656\\ \hline

\end{tabular}
\caption{Sample of Categories.}
\label{tab:cat-samples}
\end{center}
\end{table*}

\subsection{Apps Categories}

Google PlayStore allows application developers to label applications into more than one category from a set of over 50 main and sub categories. For GLARE dataset, we focused on 59 categories that we retrieved from the scraper library\ref{google-scraper} to help in crawling and scraping apps and their reviews. Apps with only one category represent the majority of the apps in GLARE dataset with 91.2\% or 9,103 apps,  followed by 673 apps with 2 categories, 184 apps with 3 categories and 20 apps with 4 categories. Since apps can have more than one category, a total of 213 different category combinations were formed. Samples of the formed application categories sets are shown in Table \ref{tab:cat-samples}.

\section{Dataset Benefits and Usage}

Plenty of NLP tasks can be explored using the GLARE dataset. Datasets for  Opinion Mining can be created by using the rating for distant supervision.  GLARE can also serve other text classification tasks, for example  classifying reviews according to their categories. Spam detection datasets can be also created from GLARE which is an important step that enhances the analysis for the reviews. Another interesting NLP task would be the study of demographic language variation as some apps attract certain demographics. Other Machine Learning (ML) tasks can be performed as well such as Applications' Ranking Prediction in App Stores.  
\medbreak
App Store reviews can work as a method of connection between software developers and their users. It can help software engineers from the step of specifying software requirements to the testing phase \cite{dkabrowski2022analysing}. 
The dataset can also be of help to software developers in analyzing their applications. It can aid them during the process of understanding their audience, and debugging errors. In addition, they can also benefit from the dataset when releasing, editing or deleting software features. It is believed that many software engineers perform maintenance measures based on the App Stores reviews   \cite{al2019app}.

\section{Conclusion and Future Work}
In this paper, we presented GLARE: Google Apps Arabic Reviews dataset. GLARE consists of 76M reviews, of which 69M are Arabic Reviews, across 9,980 Android Applications collected from the Saudi Google PlayStore. We showcased the contents of the dataset through performing EDA. We also applied feature engineering to the dataset to extract more helpful aspects from the data. As a future contribution, we are aiming to use the dataset in building a domain specific Arabic Language Model. The Language Model can be used in modeling NLP tasks that are related to software or smart devices applications' reviews. We are also planning to tackle the Arabic Aspect Based Sentiment Analysis (AABSA) task using the Arabic collected reviews, as it is one of the NLP tasks that need more exploration \cite{nassif2021deep}. Another future contribution is to create a benchmark dataset using GLARE for the tasks of Aspect Term Extraction (ATE), Aspect Category Detection (ACD) and Sentiment Analysis.

\section{Copyrights}
This work is licensed under the Creative Commons Attribution-Non-Commercial 4.0 International License (CC BY 4.0). 

\section{References}\label{reference}

\bibliographystyle{lrec2022-bib}
\bibliography{main}

\end{document}